%% file: eccv2020submission.tex
\documentclass[runningheads]{llncs}
\usepackage{graphicx}

\usepackage{comment}
\usepackage{tikz}
\usepackage{comment}
\usepackage{amsmath,amssymb} 
\usepackage{color}

\usepackage{wrapfig}
\usepackage{makecell}
\usepackage{multirow}
\usepackage{subfigure}

\DeclareMathOperator*{\argmax}{argmax}

\usepackage[font=small,labelfont=bf]{caption}

\begin{document}
\pagestyle{headings}
\mainmatter
\def\ECCVSubNumber{1}  

\title{A Machine Learning Approach to Assess Student Group Collaboration Using Individual Level Behavioral Cues} 

\author{Anirudh Som\textsuperscript{1*}, Sujeong Kim\textsuperscript{1}, Bladimir Lopez-Prado\textsuperscript{2}, Svati Dhamija\textsuperscript{1}, \\Nonye Alozie\textsuperscript{2}, Amir Tamrakar\textsuperscript{1}}
\institute{\textsuperscript{1}Center for Vision Technologies, SRI International\\ %
\textsuperscript{2}Center for Education Research and Innovation, SRI International }

\titlerunning{A Machine Learning Approach to Assess Student Group Collaboration}
\authorrunning{A. Som \emph{et al.}}

\maketitle

\begin{abstract}
K-12 classrooms consistently integrate collaboration as part of their learning experiences. However, owing to large classroom sizes, teachers do not have the time to properly assess each student and give them feedback. In this paper we propose using simple deep-learning-based machine learning models to automatically determine the overall collaboration quality of a group based on annotations of individual roles and individual level behavior of all the students in the group. We come across the following challenges when building these models: 1) Limited training data, 2) Severe class label imbalance. We address these challenges by using a controlled variant of Mixup data augmentation, a method for generating additional data samples by linearly combining different pairs of data samples and their corresponding class labels. Additionally, the label space for our problem exhibits an ordered structure. We take advantage of this fact and also explore using an ordinal-cross-entropy loss function and study its effects with and without Mixup.

\keywords{Education, Collaboration Assessment, Limited Training Data, Class Imbalance, Mixup Data Augmentation, Ordinal-Cross-Entropy Loss.}
\end{abstract}
\let\thefootnote\relax\footnotetext{*The author is currently a student at Arizona State University and this work was done while he was an intern at SRI
International.}

\input{Introduction}

\input{Related-Work}

\input{Background}

\input{Proposed-Method}

\input{Experiments}

\input{Conclusion}

\bibliographystyle{splncs04}
\bibliography{egbib}
\end{document}

%% file: Introduction.tex
\section{Introduction}

\emph{Collaboration} is identified by both the Next Generation Science Standards \cite{ngss2013next} and the Common Core State Standards \cite{daggett2010common} as a required and necessary skill for students to successfully engage in the fields of Science, Technology, Engineering and Mathematics (STEM). Most teachers in K-12 classrooms instill collaborative skills in students by using instructional methods like project-based learning \cite{krajcik2006project} or problem-based learning \cite{davidson2014boundary}. For a group of students performing a group-based collaborative task, a teacher monitors and assesses each student based on various verbal and non-verbal behavioral cues. However, due to the wide range of behavioral cues, it can often be hard for the teacher to identify specific behaviors that contribute to or detract from the collaboration effort \cite{smith2008guided,loughry2007development,taggar2001problem}. This task becomes even more difficult when several student groups need to be assessed simultaneously. 
\begin{figure}[tb!]
\centering
\includegraphics[width=0.98\linewidth]{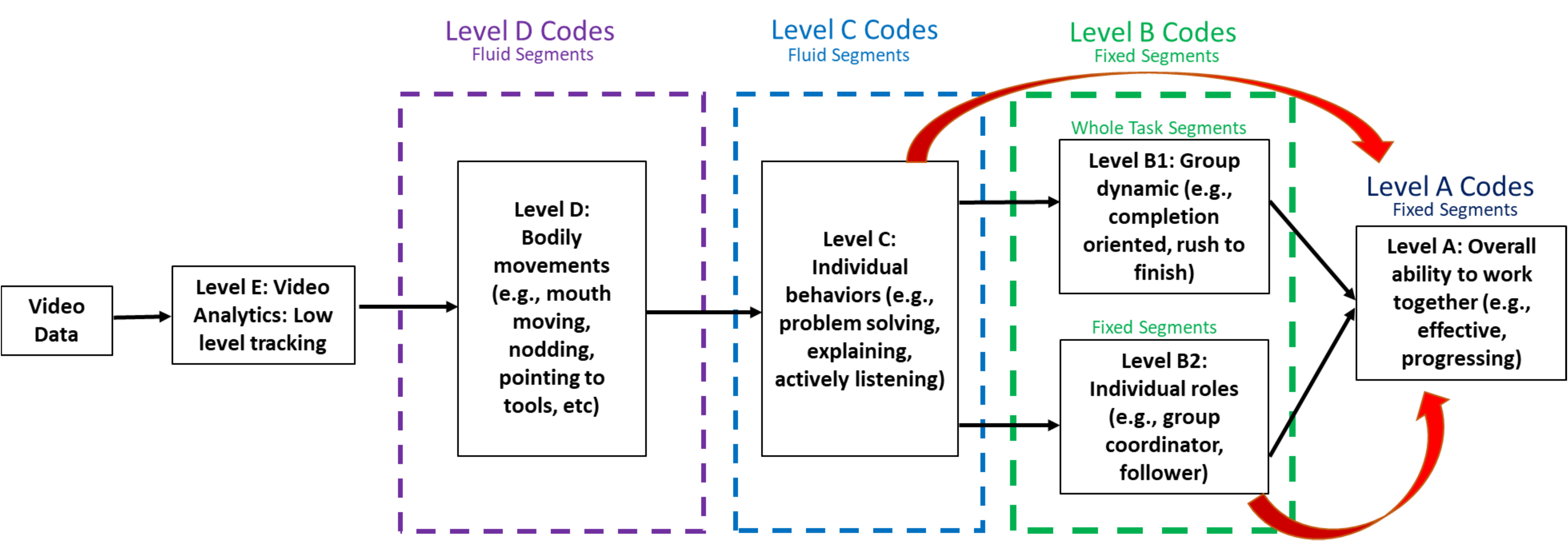}
	\caption{The collaboration assessment conceptual model. In this paper we focus on building machine learning models that map features from Level B2 $\longrightarrow$ Level A and Level C $\longrightarrow$ Level A as indicated by the red arrows. 
	}\label{figure_workflow}
\end{figure}
To better assist teachers, in our previous work we proposed an automated collaboration assessment conceptual model that provides an assessment of the collaboration quality of student groups based on behavioral communication at individual and group levels \cite{alozie2020automated,alozie2020collaboration}. 
The conceptual model illustrated in Figure \ref{figure_workflow} represents a multi-level, multi-modal integrated behavior analysis tool. The input to this model consists of Video or Audio+Video data recordings of a student group performing a collaborative task. This was done to test if visual behaviors alone could be used to estimate collaboration skills and quality. Next, low level features like facial expressions, body-pose are extracted at Level E. Information like joint attention and engagement are encoded at Level D. Level C describes complex interactions and individual behaviors. Level B is divided into two categories: Level B1 describes the overall group dynamics for a given task; Level B2 describes the changing individual roles assumed by each student in the group. Finally, Level A describes the overall collaborative quality of the group based on the information from all previous levels. This paper focuses on building machine learning models that predict a group's collaboration quality from individual roles (Level B2) and individual behaviors (Level C) of the students, indicated by red arrows in Figure \ref{figure_workflow}.

\begin{figure}[tb!]
\centering
\includegraphics[width=0.98\linewidth]{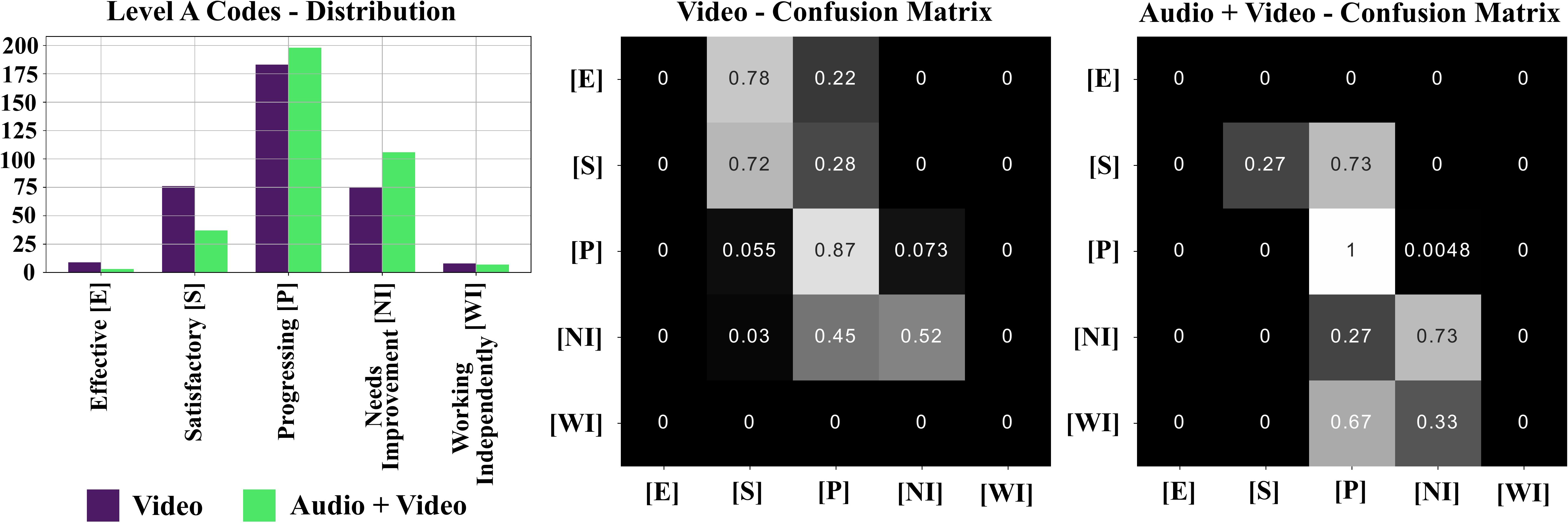}
	\caption{(left) Distribution of Level A codes which also represent the target label distribution for our classification problem. (middle, right) Aggregate confusion matrix of Multi Layer Perceptron (MLP) classification models that have been subjected to class-balancing during the training process. Even with class-balancing the MLP models are unable to overcome the bias in the training data. Note, the confusion matrix is normalized along each row, with the number in each cell representing the percentage of data samples that are classified to each class.}\label{figure_data-imbalance-issue}
\end{figure}

Deep-learning algorithms have gained increasing attention in the Educational Data Mining (EDM) community. For instance, the first papers to use deep-learning for EDM were published in 2015, and the number of publications in this field keeps growing with each year \cite{hernandez2019systematic}. Despite their growing popularity, deep-learning methods are difficult to work with under certain challenging scenarios. For example, deep-learning algorithms work best with access to large amounts of representative training data, \emph{i.e.}, data containing sufficient variations of each class label pattern. They also assume that the label distribution of the training data is approximately uniform. If either case is not satisfied then deep-learning methods tend to perform poorly at the desired task. Challenges arising due to limited and imbalanced training data is clearly depicted in Figure \ref{figure_data-imbalance-issue}. For our classification problem the label distribution appears to be similar to a bell-shaped normal distribution. As a result, for both Video and Audio+Video modality cases we have very few data samples for \emph{Effective} and \emph{Working Independently} codes, and the highest number of samples for \emph{Progressing} code. Figure \ref{figure_data-imbalance-issue} also shows the aggregate confusion matrix over all test sets after training Multi Layer Perceptron (MLP) classification models with class-balancing (\emph{i.e.}, assigning a weight to the training data sample that is inversely proportional to the number of training samples corresponding to that sample's class label). The input feature representations used were obtained from Level B2 and Level C. We observe that despite using class-balancing the predictions of the MLP model are biased towards the \emph{Progressing} code.

\textbf{Contributions:} To address the above challenges, in this paper we explore using a controlled variant of Mixup data augmentation, a simple and common approach for generating additional data samples \cite{zhang2017mixup}. Additional data samples are obtained by linearly combining different pairs of data samples and their corresponding class labels. Also note that the label space for our classification problem exhibits an ordered relationship. In addition to Mixup, we also explore the value in using an ordinal-cross-entropy loss function instead of the commonly used categorical-cross-entropy loss function.

\textbf{Outline of the paper:} Section \ref{section_related-work} discusses related work. Section \ref{section_background} provides necessary background on categorical-cross-entropy loss, ordinal-cross-entropy loss and Mixup data augmentation. Section \ref{section_proposed-method} provides description of the dataset, features extracted and the controlled variant of Mixup data augmentation. Section \ref{section_experiments} describes the experiments and results. Section \ref{section_conclusion} concludes the paper.

%% file: Related-Work.tex
\section{Related Work}\label{section_related-work}

Use of machine learning concepts for collaboration problem-solving analysis and assessment is still relatively new in the Educational Data Mining community. Reilly \emph{et al.} used Coh-Metrix indices (a natural language processing tool to measure cohesion for written and spoken texts) to train machine learning models to classify co-located participant discourse in a multi-modal learning analytics study  \cite{reilly2019predicting}. The multi-modal dataset consisted of eye-tracking, physiological and motion sensing data. They analyzed the collaboration quality between novice programmers that were instructed to program a robot to solve a series of mazes. However, they studied only two collaborative states thereby making it a binary classification problem. Huang \emph{et al.} used an unsupervised machine learning approach to discover unproductive collaborative states for the same multi-modal dataset \cite{huang2019identifying}. For input features they computed different measures for each modality. Using an unsupervised approach they were able to identify a three-state solution that showed high correlation with task performance, collaboration quality and learning gain. Kang \emph{et al.} also used an unsupervised learning approach to study the collaborative problem-solving process of middle school students. They analyzed data collected using a computer-based learning environment of student groups playing a serious game \cite{kang2019collaborative}. They used \emph{KmL}, an R package useful for applying $k$-means clustering on longitudinal data \cite{genolini2011kml}. They too identified three different states using the proposed unsupervised method. In our paper we define five different group collaboration quality states in a supervised learning setup. The above studies discuss different ways to model positive collaboration between participants in a group. For Massive Open Online Courses (MOOCs), Alexandron \emph{et al.} proposed a technique to detect cheating in the form of unauthorized collaboration using machine learning classifiers trained on data of another form of cheating (copying using multiple accounts) \cite{alexandron2020towards}.

Guo and Barmaki used a deep-learning based object detection approach for analysis of pairs of students collaborating to locate and paint specific body muscles on each other \cite{guo2019collaboration}. They used a Mask R-CNN for detecting students in video data. This is the only paper that we found that used deep-learning for collaboration assessment. They claim that close proximity of group participants and longer time taken to complete a task are indicators of good collaboration. However, they quantify participant proximity by the percentage of overlap between the student masks obtained using the Mask R-CNN. The amount of overlap can change dramatically across different view points. Also, collaboration need not necessarily be exhibited by groups that take a longer time to complete a task. In this paper, the deep-learning models are based off of the systematically designed multi-level conceptual model shown in Figure \ref{figure_workflow}. The proposed approach utilizes features at lower levels of our conceptual model but we go well beyond these and also include higher level behavior analysis as well roles taken on by students to predict the overall group collaboration quality.

We propose using Mixup augmentation, an over-sampling approach together with an ordinal cross-entropy loss function to better handle limited and imbalanced training data. Over-sampling techniques have been commonly used to make the different label categories to be approximately equal. SMOTE is one of the oldest and most widely cited over-sampling methods proposed by Chawla \emph{et al.} \cite{chawla2002smote}. The controlled variant of Mixup that we propose is very similar to their approach. However, ordinal loss functions have not received as much attention since the label space of most current classification problems of interest do not exhibit an ordered structure or relationship. We refer interested readers to the following papers that talk about ordinal loss functions for deep ordinal classification \cite{hou2016squared,beckham2017unimodal}. In this paper we propose a simple variant of the regular cross-entropy loss that takes into account the relative distance of the predicted samples from their true class label location.

%% file: Background.tex
\section{Preliminaries}\label{section_background}


\subsection{Classification Loss Functions}

Let us denote the input variables or covariates as $\mathbf{x}$, ground-truth label vector as $\mathbf{y}$ and the predicted probability distribution as $\mathbf{p}$. The cross-entropy loss \emph{a.k.a.} the categorical-cross-entropy loss function is commonly used for training deep-learning models for multi-class classification. Given a training sample $(\mathbf{x},\mathbf{y})$, the cross-entropy loss can be represented as $\text{CE}_{\mathbf{x}}(\mathbf{p},\mathbf{y}) = - \sum_{i=1}^{C} \mathbf{y}_i\log(\mathbf{p}_i)$.
Here, $C$ represents the number of classes. For a classification problem with $C$ label categories, a deep-learning model's softmax layer outputs a probability distribution vector $\mathbf{p}$ of length $C$. The $i$-th entry in $\mathbf{p}_i$ represents the predicted probability of the $i$-th class. The ground-truth label $\mathbf{y}$ is one-hot-encoded and represents a binary vector whose length is also equal to $C$. Note, $\sum_i \mathbf{y}_i = 1$ and $\sum_i \mathbf{p}_i = 1$. For an imbalanced dataset, the learnt weights of a deep-learning model will be greatly governed by the class having the most number of samples in the training set. Also, if the label space exhibits an ordinal structure, the cross-entropy loss focuses only on the predicted probability of the ground-truth class and ignores the relative distance between an incorrectly predicted data sample and its true class label. A simple variant of the cross-entropy loss that is useful for problems exhibiting an ordered label space is shown in Equation \ref{equation_ordinal-cross-entropy}.

\begin{equation}\label{equation_ordinal-cross-entropy}
    \text{OCE}_{\mathbf{x}}(\mathbf{p},\mathbf{y}) = - \left(1+w\right)\sum_{i=1}^{C} \mathbf{y}_i\log(\mathbf{p}_i), \text{ } w = \left|\argmax(\mathbf{y}) - \argmax(\mathbf{p}) \right|
\end{equation}

Here, $(1+w)$ is an additional weight that is multiplied with the regular cross-entropy loss. Within $w$, argmax returns the index of the maximum valued element in the vector and $|.|$ denotes the absolute value. During the training process, $w = 0$ for training samples that are correctly classified, with the ordinal-cross-entropy loss being the same as the cross-entropy loss. However, the ordinal-cross-entropy loss will be higher than cross-entropy loss for misclassified samples and the increase in loss is proportional to how far the samples have been misclassified from their true label locations. We later go over the benefit of using the ordinal-cross-entropy loss function in Section \ref{section_experiments-ordinality}.

\subsection{Mixup Data Augmentation}\label{subsection_mixup}

\begin{figure}[b!]
\centering
\includegraphics[width=0.999\linewidth]{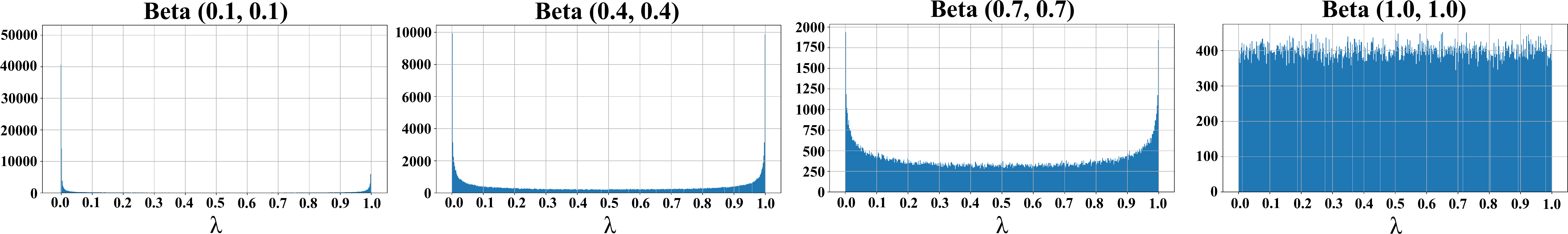}
	\caption{Beta$(\alpha,\alpha)$ distributions for $\alpha=0.1, 0.4, 0.7, 1.0$ respectively. Each Beta distribution plot  has a different y-axis range and represents a 500-bin histogram of 200000 randomly selected  $\lambda$s. Note, most $\lambda$s for Beta(0.1,0.1) are at 0 and 1.}\label{figure_variable-beta-distribution}
\end{figure}

Despite best data collection practices,  bias exists in most training datasets resulting from time or resource constraints. These biases, and the resulting performance problems of machine learning models trained on this data, are directly correlated with the problem of class imbalance. Class imbalance refers to the unequal representation or number of occurrences of different class labels. If the training data is more representative of some classes than others, then the model's predictions would systematically be worse for the under-represented classes. Conversely, with too much data or an over-representation of certain classes can skew the decision toward a particular result. Mixup is a simple data augmentation technique that can be used for imbalanced datasets \cite{zhang2017mixup}. It is used for generating additional training samples and encourages the deep-learning model to behave linearly in-between the training samples. It extends the training distribution by incorporating the prior knowledge that linear interpolations of input variables $\mathbf{x}$ should lead to linear interpolations of the corresponding target labels $\mathbf{y}$. For example, given a random pair of training samples $(\mathbf{x}^1,\mathbf{y}^1)$, $(\mathbf{x}^2,\mathbf{y}^2)$, additional samples can be obtained by linearly combining the input covariate information and the corresponding class labels. This is illustrated in Equation \ref{equation_mixup}. 

\begin{equation}\label{equation_mixup}
    \begin{split}
        \tilde{\mathbf{x}} &= \lambda\mathbf{x}^1 + (1-\lambda)\mathbf{x}^2 \\
        \tilde{\mathbf{y}} &= \lambda\mathbf{y}^1 + (1-\lambda)\mathbf{y}^2
    \end{split}
\end{equation}

Here, $(\tilde{\mathbf{x}},\tilde{\mathbf{y}})$ is the new generated sample. $\lambda\in[0,1]$ and is obtained using a Beta$(\alpha,\alpha)$ distribution with $\alpha\in(0,\infty)$. Figure \ref{figure_variable-beta-distribution} shows different Beta$(\alpha,\alpha)$ distributions for $\alpha= 0.1, 0.4, 0.7, 1.0$ respectively. If $\alpha$ approaches 0 the $\lambda$s obtained have a higher probability of being 0 or 1. If $\alpha$ approaches 1 then the Beta distribution looks more like a uniform distribution. Based on the suggestions and findings in other papers \cite{zhang2017mixup,thulasidasan2019mixup}, for our experiments we set $\alpha=0.4$. Apart from improving the classification performance on various image classification benchmarks \cite{zhang2017mixup}, Mixup also leads to better calibrated deep-learning models \cite{thulasidasan2019mixup}. This means that the predicted softmax scores of a model trained using Mixup are better indicators of the actual likelihood of a correct prediction than models trained in a regular fashion. In Section \ref{section_experiments-mixup}, we explore the benefit of using Mixup with and without ordinal-cross-entropy loss.

%% file: Proposed-Method.tex
\section{Dataset Description, Feature Extraction and Controlled Mixup Data Generation}\label{section_proposed-method}

\subsection{Dataset Description}\label{subsection_dataset_description}

Audio and video data was collected from 15 student groups across five middle schools \cite{alozie2020automated,alozie2020collaboration}. Each group was given an hour to perform 12 open-ended life science and physical science tasks that required the students to construct models of different science phenomena. 
This resulted in 15 hours of audio and video recordings. Out of the 15 groups, 13 groups had 4 students, 1 group had 3 students, and 1 group had 5 students. For Level A and Level B2, each video recording was coded by three human annotators using ELAN (an open-source annotation software) under two different modality conditions: 1) Video, 2) Audio+Video. For a given task performed by a group, each annotator first manually coded each level for the Video modality and later coded the same task for the Audio+Video modality. This was done to prevent any coding bias resulting due to the difference in modalities. A total of 117 tasks were coded by each annotator. Next, the majority vote (code) from the three coders was used to determine the ground-truth Level A code. For cases where a clear majority was not possible the median of the three codes was used as the ground-truth. We used the same code ordering depicted in Figure \ref{figure_data-imbalance-issue}. For example, if the three coders assigned \emph{Effective, Satisfactory, Progressing} for a certain task then \emph{Satisfactory} would be selected as the ground-truth label. Note, out of the 117 tasks within each modality we did not observe a majority Level A code for only 2 tasks. The distribution of the Level A target labels is shown in Figure \ref{figure_data-imbalance-issue}. For learning mappings from Level B2 $\longrightarrow$ Level A we had access to only 351 data samples (117 tasks $\times$ 3 coders) to train the machine learning models.
The protocol used for generating training-test splits is described in Section \ref{section_experiments}. In the case of Level C, each video recording was coded by just one annotator. Because of this we only had access to 117 data samples (117 tasks coded) for training the machine learning models to learn mappings from Level C $\longrightarrow$ Level A. This makes it an even more challenging classification problem. Note, the distribution of the Level A labels for this classification setting is similar to the distribution shown in Figure \ref{figure_data-imbalance-issue}, with the difference being that each label class now has just one-third of the samples.

\subsection{Level B2 and Level C Histogram Representation}\label{subsection_histogram-representations}

\begin{figure}[tb!]
	\centering
	\includegraphics[width=0.95\linewidth]{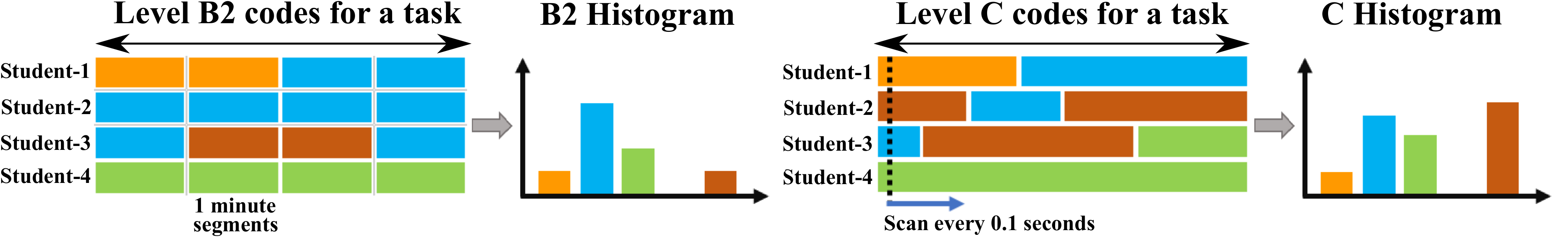}
	\caption{Histogram feature generation for Level B2 and Level C. Different colors indicate different codes assigned to each segment. Level B2 codes are represented as fixed-length 1 minute segments. Level C codes are represented as variable-length segments. A B2 histogram is generated for each task by compiling all the codes from all the students in the group. Similarly, level C histogram was generated by compiling all the codes observed after every 0.1 seconds over the duration of the task.}\label{figure_histogram-generation}
\end{figure}

For the entire length of each task, Level B2 was coded using fixed-length one minute segments and Level C was coded using variable-length segments. This is illustrated in Figure \ref{figure_histogram-generation}. 
As shown in Table \ref{table_level-b2-c-codes}, Level B2 and Level C consist of 7 codes and 23 codes respectively. Our objective in this paper is to be able to determine the overall collaboration quality of a group by summarizing all of the individual student roles and behaviors for a given task. A simple but effective way to do this is by generating histogram representations of all the codes observed in each task. Figure \ref{figure_histogram-generation} also provides a simple illustration of the histogram generation process. While it is straightforward to generate histograms for Level B2, in the case of Level C we compile all the codes observed after every 0.1 seconds for generating the histogram. Once the histogram is generated we normalize it by dividing by the total number of codes in the histogram. Normalizing the histogram in a way removes the temporal component of the task. For example, if group-1 took 10 minutes to solve a task and group-2 took 30 minutes to solve the same task, but both groups were assigned the same Level A code despite group-1 finishing the task sooner. The raw histogram representations of both these groups would look different due to the difference in number of segments coded. However, normalized histograms would make them more comparable. 

\begin{table}[tb!]
	\centering
	\caption{Coding rubric for Level B2 and Level C.}\label{table_level-b2-c-codes}
	\scalebox{0.54}{
	\begin{tabular}{ |c||c|c|c| } 
			\hline
			\textbf{Level B2 Codes} & \multicolumn{3}{c|}{\textbf{Level C Codes}} \\ 
			\hline
			Group guide/Coordinator [GG] & Talking & Recognizing/Inviting others contributions & Joking/Laughing\\
			Contributor (Active) [C] & Reading & Setting group roles and responsibilities & Playing/Horsing around/Rough housing\\
			Follower [F] & Writing & Comforting, encouraging others/Coralling & Excessive difference to authority/leader \\
			Conflict Resolver [CR] & Using/Working with materials & Agreeing & Blocking information from being shared \\
			Conflict Instigator/Disagreeable [CI] & Setting up the physical space & Off-task/Disinterested & Doing nothing/Withdrawing\\
			Off-task/Disinterested [OT] & Actively listening/Paying attention & Disagreeing & Engaging with outside environment\\
			Lone Solver [LS] & Explaining/Sharing ideas & Arguing & Waiting\\
			 & Problem solving/Negotiation & Seeking recognition/Boasting & \\
			\hline
	\end{tabular}}
\end{table}

\subsection{Controlled Mixup}


We described the simplicity and benefits of Mixup augmentation in Section \ref{subsection_mixup}. Here, we describe a controlled variant of Mixup and how it is used for our dataset. From Figure \ref{figure_data-imbalance-issue}, we know that our dataset has an imbalanced label distribution. 
Conventional Mixup selects a random pair of samples and interpolates them by a $\lambda$ that is determined using a Beta distribution. However, this results in generating samples that have the same imbalanced class distribution. We want to be able to generate a fixed number of samples for a specific category. To do this we first limit the range of $\lambda$, \emph{i.e.}, $\lambda \in [\tau,1]$, with $\tau$ being the threshold. 
Next, to generate additional samples for a specific class, we pair that class with its adjacent or neighboring classes. For example, let us use the following denotation: (primary-class, [adjacent-class-1, adjacent-class-2]), where primary-class represents the class for which we want to create additional samples; adjacent-class-1 and adjacent-class-2 represent its neighbors. We create the following pairs: (\emph{Effective}, [\emph{Satisfactory,Progressing}]),  (\emph{Satisfactory}, [\emph{Effective,Progressing}]), (\emph{Progressing}, [\emph{Satisfactory,Needs Improvement}]), (\emph{Needs Improvement}, [\emph{Progressing,Working Independently}]) and (\emph{Working Independently}, [\emph{Progressing,Needs Improvement}]). The final step consists of generating $n$ samples for the primary-class using Mixup. We do this by randomly pairing samples from the primary-class with samples from the adjacent-classes. This process is repeated $n$ times. Note that for Mixup augmentation, $\lambda$ is always multiplied with the primary-class sample and $(1-\lambda)$ is multiplied with the adjacent-class sample. For our experiments we explore the following values of $\tau$: 0.55, 0.75 and 0.95. Setting $\tau>0.5$ guarantees that the generated sample would always be dominated by the primary-class.

%% file: Experiments.tex
\section{Experiments}\label{section_experiments}

\subsubsection{Network Architecture:} We used a 5-layer Multi Layer Perceptron (MLP) model whose design was based on the MLP model described in \cite{fawaz2019deep}.
It contains the following layers: 1 input layer, 3 middle dense layers and 1 output dense layer. The normalized histogram representations discussed in Section \ref{subsection_histogram-representations} are passed as input  to the input layer. Each dense middle layer has 500 units with ReLU activation. The output dense layer has a softmax activation and the number of units is equal to 5 (total number of classes in Level A). We also used dropout layers between each layer to avoid overfitting. The dropout-rate after the input layer and after each of the three middle layers was set to 0.1, 0.2, 0.2, 0.3 respectively. We try three different types of input data: B2 histograms, C histograms and concatenating B2 and C histograms (referred to as B2+C histograms). The number of trainable parameters for B2 histogram is 507505, C histogram is 515505, B2+C histogram is 519005. Our models were developed using Keras with TensorFlow backend \cite{chollet2015keras}. We used the Adam optimizer \cite{kingma2014adam} and trained all our models for 500 epochs. The batch-size was set to one-tenth of the number of training samples during any given training-test split. We saved the best model that gave us the lowest test-loss for each training-test split.

\subsubsection{Training and Evaluation Protocol:} We adopt a round-robin leave-one-group-out cross validation protocol. This means that for each training-test split we use data from $g-1$ groups for training and the $g^\text{th}$ group is used as the test set. This process is repeated for all $g$ groups. For our experiments $g=14$ though we have histogram representations for each task performed by 15 student groups. This is because in the Audio+Video modality setting all samples corresponding to the \emph{Effective} class were found only in one group. Similarly, for the Video modality all samples corresponding to \emph{Working Independently} class were also found in just one group. Due to this reason we do not see any test samples for the \emph{Effective} class in Audio+Video and the \emph{Working Independently} class in Video in the confusion matrices shown earlier in Figure \ref{figure_data-imbalance-issue}. Note, for Level B2 $\longrightarrow$ Level A we have 351 data samples, and for Level C $\longrightarrow$ Level A we only have 117 data samples (discussed in Section \ref{subsection_dataset_description}).

\begin{figure}[b!]
	\centering
	\includegraphics[width=0.975\linewidth]{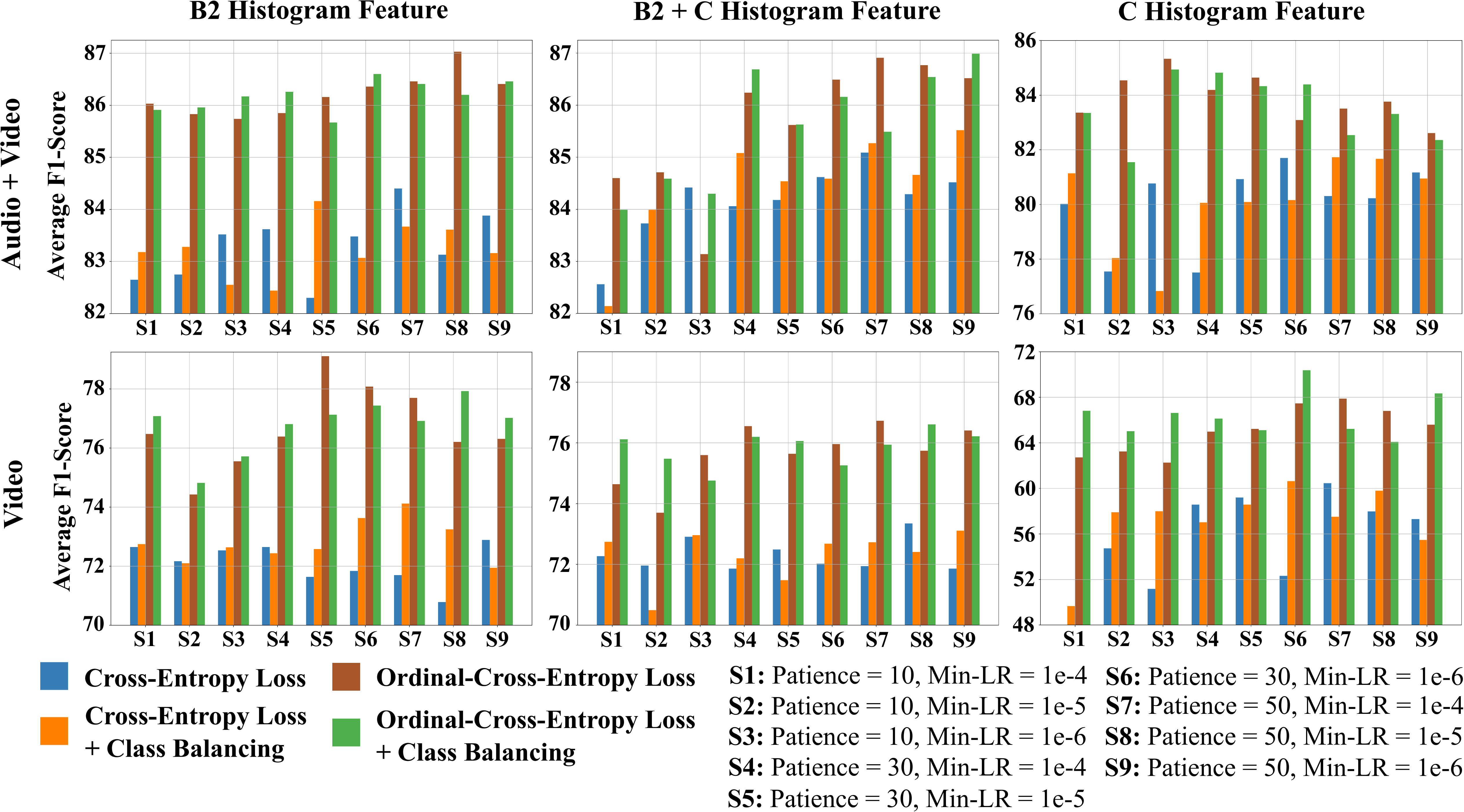}
	\caption{Comparison of the average weighted F1-Score performance between cross-entropy loss and ordinal-cross-entropy loss, with and without class balancing and under different parameter settings \textbf{S1}-\textbf{S9}.}\label{figure_strict-ordinality-effect}
\end{figure}

\subsection{Effect of Ordinal-Cross-Entropy Loss}\label{section_experiments-ordinality}

The ordinal-cross-entropy loss shown in Equation \ref{equation_ordinal-cross-entropy} takes into account the distance of the highest predicted probability from its one-hot encoded true label. This is what separates it from the regular cross-entropy loss which only focuses on the predicted probability corresponding to the ground-truth label. In this section we explore the following four variations: cross-entropy loss only, cross-entropy loss with class balancing, ordinal-cross-entropy loss only and ordinal-cross-entropy loss with class balancing. Here class balancing refers to weighting each data sample by a weight that is inversely proportional to the number of data samples corresponding to that sample's class label.

Figure \ref{figure_strict-ordinality-effect} illustrates the average weighted F1-score classification performance for the four variations under different parameter settings. We only varied the patience and minimum-learning-rate (Min-LR) parameter as we found that these two affected the classification performance the most. These parameters were used to reduce the learning-rate by a factor of 0.5 if the loss did not change after a certain number of epochs indicated by the patience parameter. Compared to the two cross-entropy-loss variants we clearly see that the two ordinal-cross-entropy loss variants help significantly improve the F1-scores across all the parameter settings. We consistently see improvements across both modality conditions and for the different histogram inputs. Using class balancing we only see marginal improvements for both loss functions. Also, the F1-scores for Video modality is always lower than the corresponding settings in Audio+Video modality. This is expected as it shows that annotations obtained using Audio+Video recordings are more cleaner and better represent the student behaviors.

\subsection{Effect of Controlled Mixup Data Augmentation}\label{section_experiments-mixup}

\begin{figure}[b!]
\centering
\includegraphics[width=0.975\linewidth]{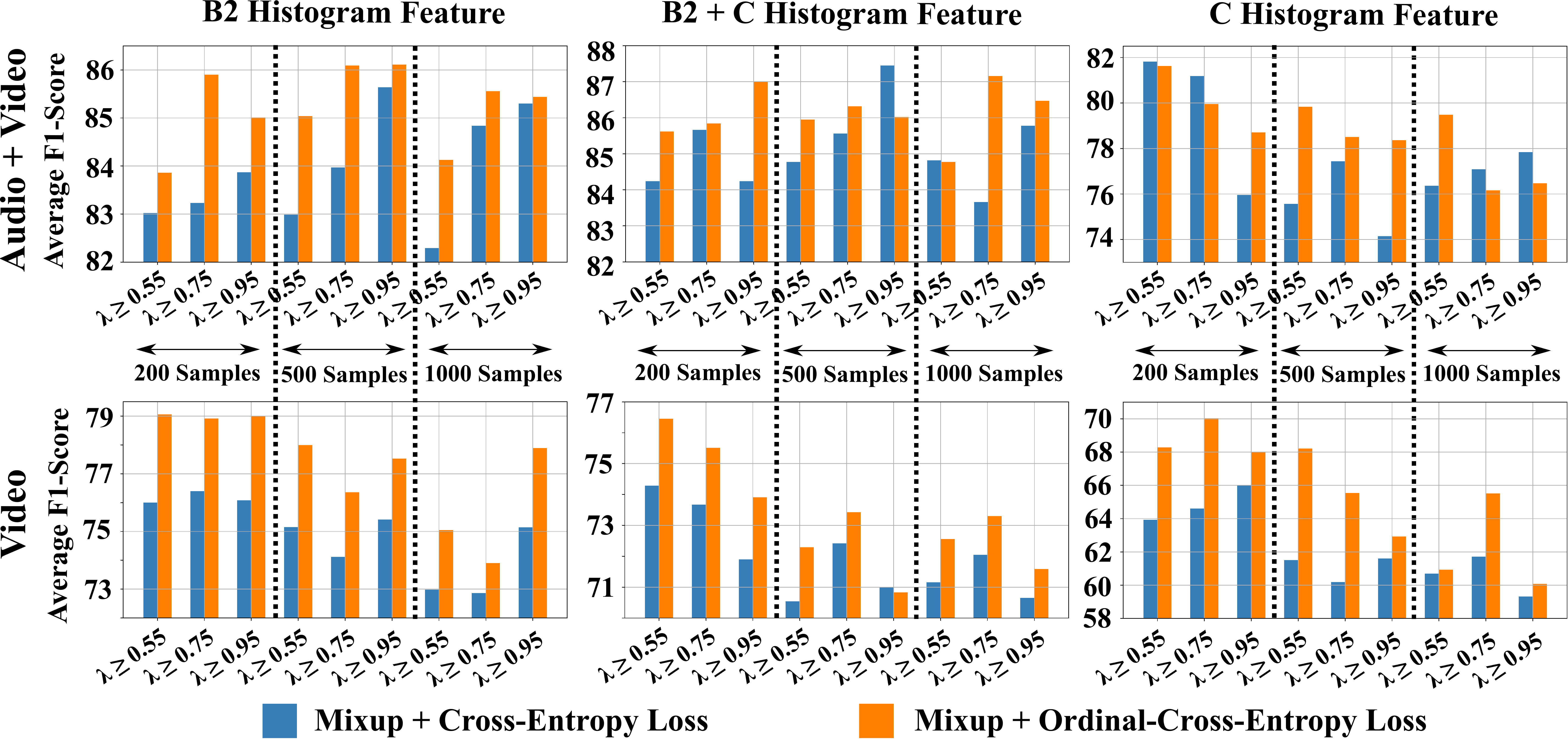}
	\caption{Comparison of the average weighted F1-Score performance of using controlled Mixup augmentation, with and without ordinal-cross-entropy loss. Here, 200, 500, 1000 samples refer to the number of samples generated per class ($n$) using controlled Mixup.}\label{figure_thresh-number-samples-mixup-effect}
\end{figure}

We use different parameter settings (listed in Figure \ref{figure_strict-ordinality-effect}) for each type of histogram input. For B2 histogram we use parameter setting S5 for both modalities. For C histogram we use parameter setting S6 for Video and S3 for Audio+Video. For B2+C histogram we use parameter setting S7 for both Video and Audio+Video. For each modality and each type of histogram input we vary threshold $\tau$ in the range 0.55, 0.75, 0.95, and also vary $n$ (the number of controlled Mixup samples generated per class) in the range of 200, 500, 1000. Note, $n =$ 200 implies that the training dataset contains 1000 samples ($n\times \text{number of classes}$). Note, this form of Mixup does not retain the original set of training samples. Figure \ref{figure_thresh-number-samples-mixup-effect} shows the effect of controlled Mixup with and without ordinal-cross-entropy loss. Across both modality settings, we observe that Mixup augmentation with ordinal-cross-entropy loss is better than Mixup with regular cross-entropy loss for all cases in B2 histogram and for most cases in C histogram and B2+C histogram. This implies that controlled Mixup and ordinal-cross-entropy loss complement each other in most cases. We also observed that having a larger $n$ does not necessarily imply better performance. For Audio+Video modality we observe F1-scores to be similar irrespective of the value of $n$. However, in the Video modality case we observe that F1-score decreases as $n$ increases. This could be attributed to the noisy nature of codes assigned by the annotators due to the lack of Audio modality. We also notice better performance using $\tau=0.75$ or $\tau=0.95$ for Audio+Video modality and $\tau=0.55$ for Video modality for B2 and B2+C histogram. However, we see the opposite effect in the case of C histogram. In the next section we will discuss two different variants of the controlled Mixup augmentation.

\begin{figure}[b!]
	\centering
	\includegraphics[width=0.975\linewidth]{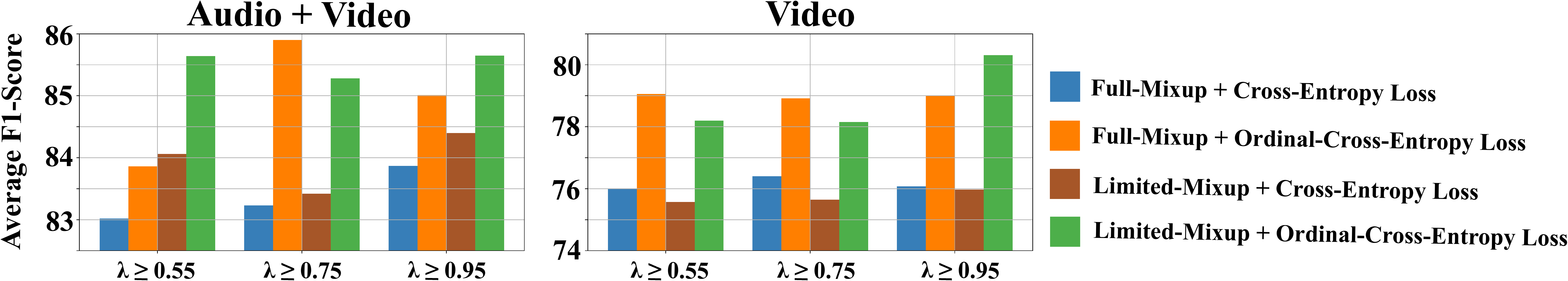}
	\caption{Full-Mixup Vs Limited Mixup evaluation using different loss functions. Average weighted F1-score shown only for B2 histogram feature input and $n=200$.}\label{figure_full-vs-limited-mixup}
\end{figure}

\subsection{Full Mixup Vs Limited Mixup}

For the controlled Mixup experiment described in the previous section, the MLP models were trained using the $n$ generated samples per class which do not retain the original set of training samples. Let us refer to this as Full-Mixup. In this section we explore training MLP models with the original set of training samples and only generate samples needed to reach $n$ samples per class using controlled Mixup. For example, let us assume that the \emph{Effective} class already has $m$ training samples, then we only compute $n-m$ samples using controlled Mixup to reach the required $n$ samples per class. This process makes sure that we always have the original set of training samples. Let us refer to this as Limited-Mixup. Figure \ref{figure_full-vs-limited-mixup} shows the average weighted F1-score comparing Full-Mixup and Limited-Mixup. We only show results for $n=200$ using B2 histogram feature as we observed similar trends in the other cases as well. We see that Full-Mixup and Limited-Mixup have similar F1-scores. This implies that we can generate the $n$ samples per class only using controlled Mixup protocol described in Section \ref{subsection_mixup} without much noticeable difference in F1-score performance.

\subsection{Additional Analysis and Discussion}

In this section we discuss in more detail the behavior of different classification models seen in the previous sections. Under each modality, Table \ref{table_best-results} shows the weighted precision, weighted recall and weighted F1-score results for the best MLP models under different experimental settings. Here, the best MLP models were decided based on the weighted F1-score since it provides a more summarized assessment by combining information seen in both precision and recall. Values in the table represent the longest bars observed in Figures \ref{figure_strict-ordinality-effect} and \ref{figure_thresh-number-samples-mixup-effect}. Note, weighted recall is equal to accuracy.  We also show results using an SVM classifier. For SVM we explored linear and different non-linear kernels with different $C$ parameter settings and only showed the best result in Table \ref{table_best-results}.

\begin{table}[tb!]
	\centering
	\caption{Weighted precision, weighted recall and weighted F1-score Mean$\pm$Std for the best MLP models under different experimental settings. The best models were selected based on the weighted F1-score. Bold values indicate the top two methods with the highest weighted precision under each modality condition.}\label{table_best-results}
	\scalebox{0.6}{
		\begin{tabular}{ |c|c|c|c|c||c|c|c| } 
			\hline
			\multirow{3}{*}{\textbf{Feature}} & \multirow{3}{*}{\textbf{Classifier}} & \multicolumn{3}{c||}{\textbf{Video}} & \multicolumn{3}{c|}{\textbf{Audio + Video}} \\ 
			\cline{3-8} &  & \textbf{\makecell{Weighted\\Precision}} & \textbf{\makecell{Weighted\\Recall}} & \textbf{\makecell{Weighted\\F1-Score}} & \textbf{\makecell{Weighted\\Precision}} & \textbf{\makecell{Weighted\\Recall}} & \textbf{\makecell{Weighted\\F1-Score}} \\
			\hline
			\multirow{11}{*}{\makecell{B2\\ Histogram}} & SVM & 74.60$\pm$11.27 & 62.67$\pm$9.42 & 63.84$\pm$11.18 & 84.45$\pm$13.43 & 73.19$\pm$16.65 & 76.92$\pm$15.39 \\
			& MLP - Cross-Entropy Loss & 76.90$\pm$12.91 & 73.95$\pm$11.02 & 72.89$\pm$13.22 & 83.72$\pm$16.50 & 86.42$\pm$10.44 & 84.40$\pm$13.85 \\
			& \makecell{MLP - Cross-Entropy Loss \\ + Class-Balancing} & 77.08$\pm$13.03 & 73.84$\pm$13.27 & 74.12$\pm$13.59 & 83.93$\pm$17.89 & 85.29$\pm$14.37 & 84.16$\pm$16.23 \\
			& MLP - Ordinal-Cross-Entropy Loss & 81.51$\pm$13.44 & 79.09$\pm$13.62 & 79.11$\pm$13.96 & 86.96$\pm$14.56 & 88.78$\pm$10.36 & 87.03$\pm$13.16 \\
			& \makecell{MLP - Ordinal-Cross-Entropy Loss \\ + Class-Balancing} & 80.78$\pm$14.12 & 78.70$\pm$11.98 & 77.93$\pm$14.05 & 86.73$\pm$14.43 & 88.20$\pm$9.66 & 86.60$\pm$12.54 \\
			& \makecell{\textbf{MLP - Cross-Entropy Loss} \\ + \textbf{Mixup}} & \textbf{81.61$\pm$12.81} & \textbf{73.56$\pm$10.31} & \textbf{76.40$\pm$11.00} & 88.51$\pm$12.32 & 83.58$\pm$14.14 & 85.64$\pm$13.23 \\
			& \makecell{\textbf{MLP - Ordinal-Cross-Entropy Loss} \\ \textbf{+ Mixup}} & \textbf{83.30$\pm$10.06} & \textbf{76.57$\pm$9.42} & \textbf{79.06$\pm$9.66} & 89.59$\pm$10.15 & 84.93$\pm$13.20 & 86.09$\pm$12.94 \\
			\hline
			
			\multirow{11}{*}{\makecell{C\\ Histogram}} & SVM & 59.27$\pm$27.00 & 42.76$\pm$20.69 & 46.85$\pm$22.26 & 72.33$\pm$20.33 & 60.15$\pm$19.45 & 63.25$\pm$17.96 \\
			& MLP - Cross-Entropy Loss & 63.24$\pm$20.78 & 65.73$\pm$16.34 & 60.46$\pm$17.57 & 81.15$\pm$16.90 & 84.16$\pm$11.67 & 81.70$\pm$14.41 \\
			& \makecell{MLP - Cross-Entropy Loss \\ + Class-Balancing} & 63.82$\pm$22.08 & 64.77$\pm$18.51 & 60.64$\pm$19.89 & 80.44$\pm$18.11 & 84.88$\pm$11.70 & 81.67$\pm$15.06 \\
			& MLP - Ordinal-Cross-Entropy Loss & 68.16$\pm$27.13 & 72.59$\pm$17.88 & 67.88$\pm$23.01 & 86.05$\pm$14.11 & 86.90$\pm$11.43 & 85.33$\pm$13.07 \\
			& \makecell{MLP - Ordinal-Cross-Entropy Loss \\ + Class-Balancing} & 71.74$\pm$24.34 & 74.10$\pm$16.75 & 70.37$\pm$20.94 & 85.24$\pm$13.54 & 86.11$\pm$11.65 & 84.94$\pm$12.52 \\
			& \makecell{MLP - Cross-Entropy Loss \\ + Mixup} & 72.27$\pm$23.29 & 64.45$\pm$19.55 & 66.02$\pm$20.35 & 84.25$\pm$13.78 & 81.91$\pm$13.68 & 81.82$\pm$13.93 \\
			& \makecell{MLP - Ordinal-Cross-Entropy Loss \\ + Mixup} & 75.11$\pm$21.63 & 69.54$\pm$18.64 & 70.03$\pm$20.01 & 82.94$\pm$14.63 & 81.91$\pm$14.68 & 81.63$\pm$14.46 \\
			\hline
			
			\multirow{11}{*}{\makecell{B2+C\\ Histogram}} & SVM & 72.49$\pm$15.35 & 61.89$\pm$13.21 & 64.95$\pm$14.15 & 82.32$\pm$16.53 & 73.32$\pm$15.27 & 76.65$\pm$15.65 \\
			& MLP - Cross-Entropy Loss & 76.15$\pm$15.81 & 74.59$\pm$15.02 & 73.35$\pm$16.08 & 83.38$\pm$19.42 & 87.75$\pm$14.68 & 85.09$\pm$17.12 \\
			& \makecell{MLP - Cross-Entropy Loss \\ + Class-Balancing} & 75.75$\pm$17.23 & 73.81$\pm$16.50 & 73.11$\pm$17.17 & 84.71$\pm$16.57 & 88.68$\pm$11.04 & 85.52$\pm$15.01 \\
			& MLP - Ordinal-Cross-Entropy Loss & 78.05$\pm$17.94 & 77.88$\pm$16.16 & 76.73$\pm$17.65 & 85.51$\pm$17.28 & 89.25$\pm$12.19 & 86.91$\pm$14.99 \\
			& \makecell{MLP - Ordinal-Cross-Entropy Loss \\ + Class-Balancing} & 78.10$\pm$17.70 & 77.33$\pm$17.02 & 76.61$\pm$17.96 & 86.90$\pm$15.83 & 88.82$\pm$11.50 & 86.99$\pm$14.15 \\
			& \makecell{\textbf{MLP - Cross-Entropy Loss} \\ + \textbf{Mixup}} & 77.99$\pm$17.42 & 72.86$\pm$14.32 & 74.29$\pm$16.08 & \textbf{90.48$\pm$11.20} & \textbf{86.57$\pm$14.07} & \textbf{87.45$\pm$13.65} \\
			& \makecell{\textbf{MLP - Ordinal-Cross-Entropy Loss} \\ + \textbf{Mixup}} & 77.92$\pm$16.66 & 75.82$\pm$15.27 & 76.45$\pm$16.02 & \textbf{90.05$\pm$10.80} & \textbf{85.91$\pm$14.00} & \textbf{87.01$\pm$13.18} \\
			\hline
	\end{tabular}}
	
\end{table}

For both modalities and for each type of histogram input, if we focus only on the weighted F1-scores we notice that there is little or no improvement as we go towards incorporating controlled Mixup and ordinal-cross-entropy loss. For this reason we also show the corresponding weighted precision and weighted recall values. We observe that the average weighted precision increases and the standard-deviation of the weighted precision decreases as we go towards the proposed approach. For an imbalanced classification problem the objective is to be able to predict more true positives. Thus a higher precision indicates more true positives as it does not consider any false negatives in its calculation. The bold values in Table \ref{table_best-results} indicate the top two methods with highest weighted precision values in each modality. We find that the Cross-Entropy loss + Mixup and Ordinal-Cross-Entropy loss + Mixup methods show the highest weighted Precision using the B2 histogram input in the Video modality and the B2+C histogram input in the Audio+Video modality.

\begin{figure}[tb!]
\centering
\includegraphics[width=0.99\linewidth]{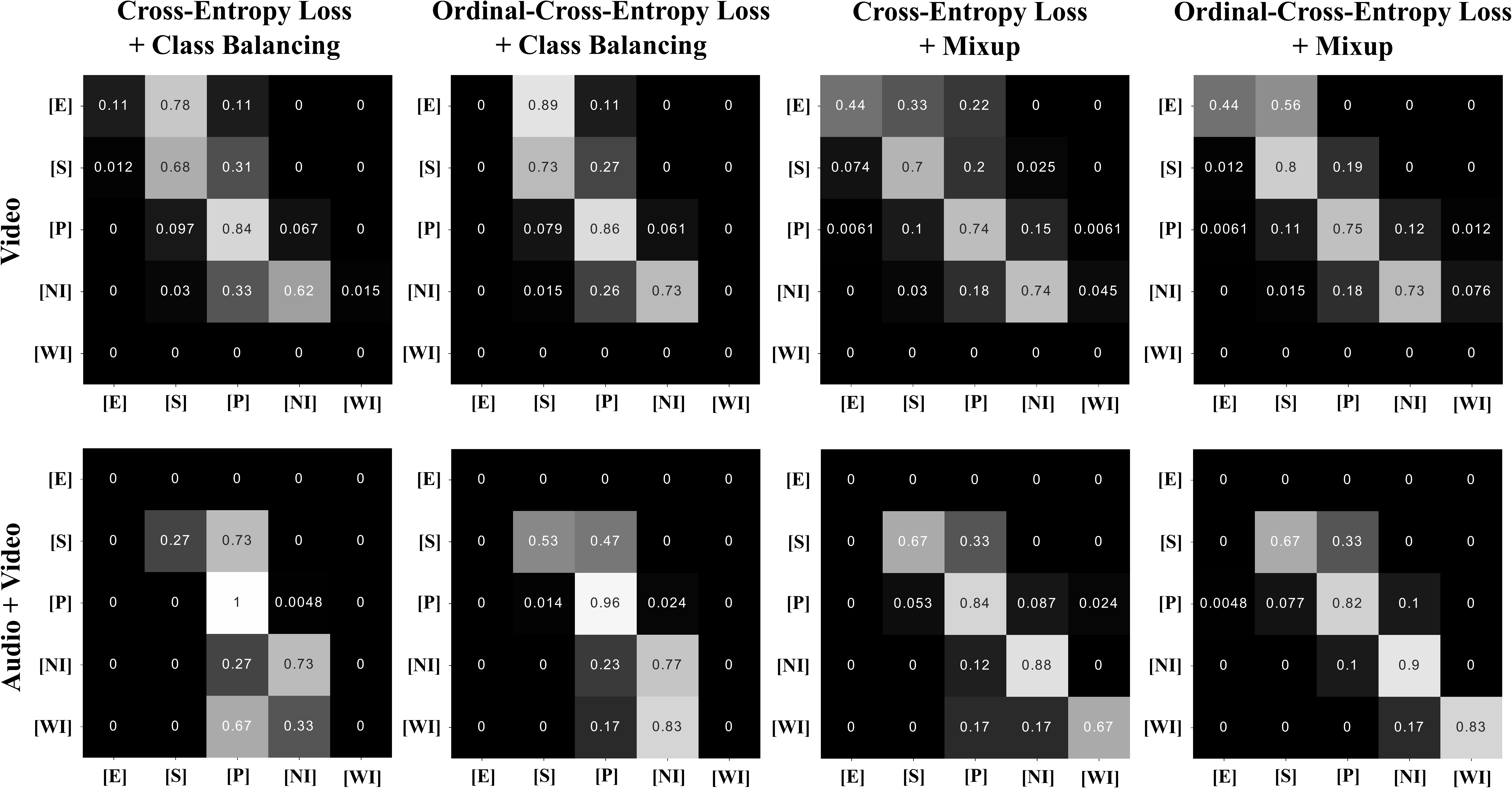}
	\caption{Aggregate confusion matrix illustrations of the MLP classification model under different experimental conditions. The confusion matrix for each method corresponds to the best MLP model described in Table \ref{table_best-results}. The confusion matrices are normalized along each row. Note, the number in each cell represents the percentage of samples classified to each class.}\label{figure_confusion-matrix}
\end{figure}

The higher weighted precision is better illustrated using the confusion matrices shown in Figure \ref{figure_confusion-matrix}. Here, we show confusion matrices for Video modality using the B2 histogram features and for Audio+Video modality using the B2+C histogram, as these showed the best weighted precision values in Table \ref{table_best-results}. As seen earlier in Section \ref{section_experiments-ordinality}, ordinal-cross-entropy loss did show significant improvements in terms of weighted F1-score. However, even with class balancing we notice that the best MLP model is still biased towards the class with the most training samples. If we look at the controlled Mixup variants with either cross-entropy loss or ordinal-cross-entropy loss we notice a better diagonal structure in the confusion matrix, indicating more number of true positives. Note, we do not see any test samples for the \emph{Effective} class in Audio+Video and the \emph{Working Independently} class in Video in the confusion matrices. Between Cross-Entropy loss + Mixup and Ordinal-Cross-Entropy loss + Mixup, we notice that ordinal-cross-entropy loss helps minimize the spread of test sample prediction only to the nearest neighboring classes.


%% file: Conclusion.tex
\section{Conclusion}\label{section_conclusion}

In this paper we built simple machine learning models to determine the overall collaboration quality of a student group based on the summary of individual roles and individual behaviors exhibited by each student. We come across challenges like limited training data and severe class imbalance when building these models. To address these challenges we proposed using an ordinal-cross-entropy loss function together with a controlled variation of Mixup data augmentation. Ordinal-cross-entropy loss is different from the regular categorical cross-entropy loss as it takes into account how far the training samples have been classified from their true label locations. We proposed a controlled variant of Mixup allowing us to generate a desired number of data samples for each label category for our problem. Through various experiments we studied the behavior of different machine learning models under different experimental conditions and realized the benefit of using ordinal-cross-entropy loss with Mixup. For future work, we would like to explore building machine learning models that learn mappings across the other levels described in Figure \ref{figure_workflow} and also explore the temporal nature of the annotation segments.